\documentclass[lettersize,journal]{IEEEtran}
\usepackage{amsmath,amsfonts}
\usepackage{algorithmic}
\usepackage{algorithm}
\usepackage{array}
\usepackage[caption=false,font=normalsize,labelfont=sf,textfont=sf]{subfig}
\usepackage{textcomp}
\usepackage{stfloats}
\usepackage{url}
\usepackage{verbatim}
\usepackage{graphicx}
\usepackage{cite}

\usepackage{tikz}
\usetikzlibrary{positioning, shapes.geometric, shapes.multipart, arrows.meta, shadows, calc}
\usepackage{amsmath}
\usepackage{xcolor}
\usepackage{siunitx}
\usepackage{comment}
\usepackage{booktabs}
\usepackage{multirow}
\usepackage{algorithm}
\usepackage{algorithmic}
\usepackage{newfloat}
\usepackage{listings}
\begin{document}

\title{A Spatio-Temporal Deep Learning Approach For High-Resolution Gridded Monsoon Prediction}


\author{
    Parashjyoti Borah\textsuperscript{1},
    Sanghamitra Sarkar\textsuperscript{1},
    Ranjan Phukan\textsuperscript{1}\\
    \textsuperscript{1}Department of Computer Science and Engineering,\\
    Indian Institute of Information Technology Guwahati, India\\
    parashjyoti@iiitg.ac.in, sanghamitra.sarkar@iiitg.ac.in, ranjan.phukan@iiitg.ac.in
}



\maketitle

\begin{abstract}
    The Indian Summer Monsoon (ISM) is a critical climate phenomenon, fundamentally impacting the agriculture, economy, and water security of over a billion people. Traditional long-range forecasting, whether statistical or dynamical, has predominantly focused on predicting a single, spatially-averaged seasonal value, lacking the spatial detail essential for regional-level resource management. To address this gap, we introduce a novel deep learning framework that reframes gridded monsoon prediction as a spatio-temporal computer vision task. We treat multi-variable, pre-monsoon atmospheric and oceanic fields as a sequence of multi-channel images, effectively creating a video-like input tensor. Using 85 years of ERA5 reanalysis data for predictors and IMD rainfall data for targets, we employ a Convolutional Neural Network (CNN)-based architecture to learn the complex mapping from the five-month pre-monsoon period (January-May) to a high-resolution gridded rainfall pattern for the subsequent monsoon season. Our framework successfully produces distinct forecasts for each of the four monsoon months (June-September) as well as the total seasonal average, demonstrating its utility for both intra-seasonal and seasonal outlooks.
\end{abstract}


\section{Introduction}
    \IEEEPARstart{T}{he} Indian Summer Monsoon (ISM) is a critical component of the global climate system, delivering about 78\% of the annual rainfall to the Indian subcontinent and profoundly impacting its agriculture, economy, and the livelihood of over a billion people \cite{kumar2024critical}. Given its societal importance, the long-range forecasting of the monsoon has been a subject of intense scientific pursuit for over a century. Traditionally, these forecasting efforts have been dominated by two primary paradigms. The first is statistical methods, pioneered by the seminal work of Sir Gilbert Walker \cite{Walker1923, Walker1924}, which uses the historical relationship between the Indian Summer Monsoon Rainfall (ISMR) and global atmosphere–ocean parameters \cite{gowariker1989parametric,rajeevan2000new,pai2006long,rajeevan2007new}. The second, and more modern, approach relies on dynamical models, specifically complex General Circulation Models (GCMs) or Numerical Weather Prediction (NWP) systems, which simulate the physics of the ocean and atmosphere to predict future states \cite{gadgil2011seasonal,george2016indian}. While dynamical models are the backbone of modern operational forecasting, they are computationally expensive and often suffer from systematic biases that limit their predictive skill.

    Recent advancements in Artificial Intelligence (AI) and the availability of extensive climate datasets, such as ERA5 reanalysis, have ushered in a third paradigm: data-driven forecasting. Machine Learning (ML) and Deep Learning (DL) models excel at identifying complex, non-linear patterns and teleconnections directly from observational data, bypassing the constraints of explicit physical equations. Techniques such as Support Vector Machines (SVMs), Random Forests, and Artificial Neural Networks (ANNs) have been applied to forecast spatially averaged monsoon indices, often outperforming traditional methods \cite{Ham2019,saha2017deep,monir2023spatiotemporal}. Advanced DL architectures, including Convolutional Neural Networks (CNNs) and Convolutional Long Short-Term Memory (ConvLSTM) networks, have shown promise in capturing spatio-temporal patterns for short- to medium-range precipitation forecasting \cite{w11050977,Guo2024,chattopadhyay2020analog}. Additionally, transformer-based models, such as Vision Transformers and FourCastNet, have demonstrated potential for global weather forecasting by leveraging attention mechanisms to model long-range dependencies \cite{pathak2022fourcastnet,nguyen2023climaxfoundationmodelweather}.
    
    Despite these advances, a significant gap persists in both operational and research-oriented forecasting: the prediction of high-resolution, long-range monsoon rainfall. Most existing models, whether statistical, dynamical, or ML-based, primarily focus on short- to medium-range forecasts \cite{kim2017deeprain,dash2024deep} or spatially aggregated value (e.g., the all-India seasonal rainfall anomaly) \cite{gadgil2012monsoon, w11050977}, leaving a gap in long-range, high-resolution, period-specific monsoon predictions. While some statistical downscaling techniques can produce gridded forecasts, they typically rely on training independent models for each grid point or station \cite{tripathi2006downscaling}. This approach is not only computationally inefficient at scale but, more critically, it cannot inherently learn the complex spatial interdependencies that define a monsoon system. Consequently, these predictions lack the holistic spatial detail required for effective regional-level planning in agriculture and water resource management. In this work, we address this critical challenge by proposing a novel framework that reframes gridded monsoon prediction as a computer vision task. We treat the multi-variable, pre-monsoon atmospheric and oceanic fields as a sequence of multi-channel images, effectively creating a video-like input tensor. We refer to the “pre-monsoon” period as the five months preceding the Indian Summer Monsoon (January to May), rather than the conventional March–May definition, to better capture the evolving large-scale atmospheric and oceanic signals. This structure allows us to leverage powerful deep learning architectures designed for spatio-temporal feature extraction to predict a high-resolution, gridded pattern of the subsequent monsoon rainfall.

\section{Related Work}
\label{sec:related_work}
    The challenge of long-range monsoon forecasting has been approached from three distinct perspectives: statistical modeling, dynamical simulation, and more recently, machine learning-based approaches. We review related work in this section to contextualize our study.

    \subsection{Statistical and Dynamical Forecasting}
        Forecasting the Indian Summer Monsoon has long been a scientific challenge due to its complex and nonlinear nature. Traditionally, two core methodologies have guided operational and research efforts in statistical forecasting, which leverages empirical relationships from historical data, and dynamical forecasting, which simulates atmospheric processes using physical models.\\
        
        \subsubsection{Statistical Forecasting} Statistical methods for forecasting the Indian Summer Monsoon Rainfall (ISMR) identify predictors by analyzing historical data to find variables with strong, statistically significant correlations to monsoon variability. This approach originated after the 1877–78 famine, when H. F. Blanford of IMD linked Himalayan snow cover to monsoon rainfall \cite{blanford1884connection}. Sir John Elliot later incorporated additional predictors, including atmospheric conditions over the Indian Ocean and Australia and local weather patterns in India \cite{rao2019seasonal,thapliyal1987prediction}. A major advancement came from Sir Gilbert Walker, who introduced the concept of global pressure oscillations, especially the Southern Oscillation, leading to the development of region-specific multiple regression models \cite{Walker1923,Walker1924}. This empirical framework formed the basis of operational forecasting until the late 1980s. After a major forecast failure in 2002, IMD critically re-evaluated the 16-parameter models \cite{gowariker1989parametric,gowariker1991power} and introduced two new models with 8 and 10 predictors in 2003, along with a revised two-stage forecast system \cite{rajeevan2004imd}. Further advancements were made through the incorporation of improved statistical methodologies and model refinement techniques, as demonstrated by \cite{rajeevan2007new}. While efficient, statistical models often fall short in capturing the nonlinear dynamics of the climate system.
        
        \subsubsection{Dynamical Forecasting} To overcome limitations of statistical models, operational forecasting has increasingly adopted dynamical models, especially coupled General Circulation Models (GCMs), which simulate climate evolution based on physical laws. Initialized with observed conditions, these models can predict key drivers like El Niño-Southern Oscillation (ENSO), Indian Ocean Dipole (IOD), and monsoon circulation, enabling forecasts beyond historical correlations. A major advancement is the use of ensemble systems, where multiple simulations account for uncertainty. Multi-model ensembles (MMEs) such as NMME, EUROSIP and India’s CFSv2-based Monsoon Mission system enhance reliability by averaging model errors and providing probabilistic forecasts \cite{saha2014ncep,jain2024mmcfs}. These are now standard at major centers like IMD, National Oceanic and Atmospheric Administration (NOAA), European Centre for Medium-Range Weather Forecasts (ECMWF) and the United Kingdom Meteorological Office (UKMO). While offering a physically consistent framework, dynamical models are computationally demanding and exhibit biases in tropical rainfall and SST simulation. They also face the spring predictability barrier, limiting skill for forecasts initialized in boreal spring \cite{jin2008current}. Still, they mark a substantial improvement over empirical approaches.       
    
    \subsection{Machine Learning in Climate Prediction}
        The proliferation of large, high-resolution climate datasets has spurred the application of machine learning (ML) as a third paradigm. This research can be broadly categorized into two main areas: index prediction and gridded forecasting.
        
        \subsubsection{Index Prediction} A significant body of work has focused on using ML to predict spatially-averaged climate indices. This includes classical ML models like Support Vector Machines and Random Forests, as well as deep neural networks, for forecasting the all-India summer monsoon rainfall (AISMR) index \cite{chattopadhyay2016elucidating,kundu2017future,das2017random,das2024comparative,singh2013indian}. A landmark study by \cite{Ham2019} demonstrated that a CNN could predict the ENSO index with high accuracy up to 1.5 years in advance by analyzing historical sea surface temperature data. These works powerfully demonstrate the ability of deep learning to learn from spatio-temporal data, but are ultimately limited to predicting a single scalar value, which lacks the regional detail needed for practical applications.
        
        \subsubsection{Gridded Spatio-Temporal Forecasting} More recently, deep learning has been applied to the more challenging task of producing gridded (map-based) forecasts. In short-term weather prediction (nowcasting), U-Net architectures have proven highly effective for predicting precipitation patterns hours in advance \cite{Shi2015, Agrawal2019}. For medium-range global weather forecasting, data-driven models like FourCastNet \cite{pathak2022fourcastnet} and DeepMind's GraphCast \cite{lam2023graphcast} have shown remarkable skill, outperforming traditional NWP models on many metrics. FourCastNet utilises Adaptive Fourier Neural Operators (AFNOs) to execute spectral transformations, therefore effectively capturing global spatial dependencies in atmospheric fields \cite{pathak2022fourcastnetglobaldatadrivenhighresolution}. The architecture comprises repetitive AFNO blocks that calculate spectral coefficients, perform nonlinear activations, and execute inverse transforms, facilitating rapid global forecasts utilising multivariable inputs from the ERA5 reanalysis dataset at a 0.25° resolution, while achieving several orders of magnitude speed-up compared to numerical weather prediction systems \cite{10.1145/3592979.3593412}. whereas,GraphCast employs a Graph Neural Network (GNN) to map the Earth on an icosahedral grid and learn spatial correlations through message transmission between nodes \cite{lam2023learning}. Trained on reanalysis data, GraphCast generates medium-range global forecasts of multiple weather variables for lead times up to 10 days at approximately 0.25° resolution. These architectures demonstrate that data-driven, globally coherent models can rival or surpass operational forecasting systems in both skill and computational efficiency.
        
        However, applying these successes to long-range seasonal climate forecasting remains a frontier. While some studies have explored CNNs for predicting gridded seasonal anomalies of variables like temperature \cite{feng2022predictability}, they often do not tackle the high variability of precipitation or the specific complexities of the monsoon. Furthermore, many approaches simplify the input to a single variable (e.g., SST) or a static 2D image, failing to leverage the rich, dynamic, multi-variable evolution of the pre-monsoon climate system \cite{CHOI2023105262}.\\
    
        Our work is positioned at the intersection of these research threads and addresses a clear gap. While previous works have proven DL's utility for index prediction and short-term weather forecasting, the challenge of high-resolution, long-range monsoon forecasting remains largely unmet.

\section{A Deep Learning Framework for Gridded Monsoon Forecasting}
\label{sec:methodology}

    This study is built upon a framework constructed to frame long-range monsoon prediction as a supervised learning problem. We transform raw, large-scale climate records into highly structured tensors, mapping spatio-temporal pre-monsoon conditions (predictors) to subsequent gridded rainfall patterns (targets). The following subsections detail the three core components of our work: the end-to-end pipeline for dataset construction, the design of a deep learning architecture, and the protocol for model training and evaluation.
    
    \subsection{Dataset Construction and Preprocessing}
    \label{ssec:dataset}
        Our approach frames monsoon forecasting as a sequence-to-frame prediction task, where a sequence of past atmospheric states is used to predict a future rainfall map. The predictor variables consist of spatio-temporal sequences of key atmospheric fields derived from the ERA5 reanalysis dataset \cite{hersbach2020era5,ecmwf_era5}\footnote{\url{https://www.ecmwf.int/en/forecasts/dataset/ecmwf-reanalysis-v5}}.
        The target variable comprises corresponding gridded average daily rainfall frames, obtained from the India Meteorological Department (IMD) \cite{imd_rainfall_data}\footnote{\url{https://www.imdpune.gov.in/cmpg/Griddata/Rainfall_1_NetCDF.html}}.
        The complete pipeline is described below, covering data acquisition, spatio-temporal alignment, normalization, downsampling, and data augmentation to ensure data integrity and suitability for deep learning models.

        \subsubsection{Predictor Dataset: Pre-Monsoon Spatio-Temporal Fields}
        \label{sssec:predictor}
            The predictor dataset was constructed using the ERA5 reanalysis product \cite{hersbach2020era5}, which provides a globally complete and consistent climate record. Raw daily data at a native 0.25$^{\circ}$ $\times$ 0.25$^{\circ}$ spatial resolution was sourced for the period 1940 to 2024, yielding an 85-year corpus.
            
            A specific Region of Interest (ROI) was defined spanning 20$^{\circ}$S to 45$^{\circ}$N latitude and 30$^{\circ}$E to 165$^{\circ}$E longitude. This domain was strategically selected to encapsulate the key drivers of the Indian Summer Monsoon, including the Indian Ocean basin, the Tibetan Plateau, the Maritime Continent, and parts of the Western Pacific.
            
            To capture the evolution of pre-monsoon conditions, we processed daily data from January 1st to May 31st of each year. This pre-monsoon window was temporally downsampled by computing fortnightly means. The process generates a sequence of 11 temporal frames for each year; while the first ten frames are true 14-day averages, the final frame covers the remaining 11 or 12 days of the period. This sequence provides a temporally evolving representation of the atmospheric system's state across the pre-monsoon period.
            
            The model input is constructed from two categories of variables, chosen to provide a comprehensive view of the coupled ocean-atmosphere system \cite{gowariker1989parametric,rajeevan2004imd,rajeevan2007new}. The first category includes five key atmospheric variables: geopotential height (\textit{z}), specific humidity (\textit{q}), air temperature (\textit{t}), and the zonal (\textit{u}) and meridional (\textit{v}) components of wind. These were sampled at four standard pressure levels critical for monsoon dynamics: 850, 700, 500, and 200 hPa. The second category consists of five single-level surface and total-column variables: Sea Surface Temperature (SST), Mean Sea Level Pressure (MSLP), 2-meter air temperature (t2m), total column water vapor (tcwv), and total precipitation (tp). All the daily data corresponds to a single hourly snapshot at 00:00 UTC. This combination results in (5 variables $\times$ 4 levels) + 5 single-level variables, yielding a total of 25 distinct two-dimensional gridded fields. Each of these 25 fields was treated as a separate channel, analogous to the RGB channels in a computer vision task. The final structure for the predictor data is a 5-dimensional tensor with a shape of \texttt{(Years, Fortnights, Latitude, Longitude, Channels)}, which for our dataset is specifically \texttt{(85, 11, 261, 541, 25)}.
            
            The 85-year dataset was systematically divided into training and testing sets to prevent chronological data leakage. Years starting from 1944 at 4-year intervals (i.e., 1944, 1948, ..., 2024) were allocated to the test set (21 years), with the remaining 64 years forming the training set.
            
            Each of the 25 predictor channels was independently scaled to a (0, 1) range using Min-Max normalization. The scaling parameters (minimum and maximum values) for each channel were computed solely from the training set data to prevent information leakage. The test set was subsequently scaled using these stored parameters, and any resulting values outside the [0, 1] range were clipped. A crucial imputation step was performed for the SST channel, where land-based `NaN` values were set to a distinct value of -1.0 after normalization, allowing the model to explicitly identify land areas.

            A spatial downsampling step was performed on the predictor dataset to reduce computational and memory demands and to encourage the models to learn from larger-scale atmospheric patterns. Each of the 25 predictor channels was downsampled in both spatial dimensions by a factor of 3 using average pooling with a 3×3 window and stride of (3, 3). This procedure reduced the spatial resolution of the predictor fields from the native \(0.25^\circ \times 0.25^\circ\) grid to an effective resolution of \(0.75^\circ \times 0.75^\circ\). To ensure the dimensions were perfectly divisible by the factor, the spatial grid was first trimmed from its original size of (261, 541) to (261, 540). This resulted in final downsampled dimensions of (87, 180) for latitude and longitude, respectively. This downsampling was applied to all samples in both the training and testing sets after the normalization step.

            To address the limited size of the training set (64 years) and enhance model robustness, we implemented a bespoke sliding window-based augmentation scheme, applied exclusively to the training data. This strategy generates new training samples by systematically altering spatial regions within the input predictors. For each original training sample, a sliding window with a shape of (40, 60) (latitude, longitude) moves across the spatial domain with a stride of (26, 40). At each window position, two types of augmented samples are created:
            \begin{enumerate}
                \item \textit{Inclusive Patch Sample:} A new sample is generated containing only the data within the current window patch, while the surrounding area is masked with a constant \texttt{fill\_value} of -1.0. This forces the model to learn the predictive importance of localized features in isolation.
                \item \textit{Occlusive Patch Sample:} A copy of the original sample is made, but the data within the current window patch is occluded with the same \texttt{fill\_value}. This encourages the model to learn from contextual information and prevents over-reliance on any single sub-region.
            \end{enumerate}
            By retaining the original, un-augmented data, the augmentation process expands the training corpus significantly from 64 to 1088 samples. Crucially, for all augmented versions, the target rainfall map remains unchanged. The final augmented dataset is then randomly shuffled to ensure unbiased model training. To ensure a fair and reproducible comparison, the same shuffled index sequence is used for all the experiments.
            
        \subsubsection{Target Variable: Gridded Monsoon Rainfall}
        \label{sssec:target}
            The target variable for the prediction task is monsoon rainfall over the Indian subcontinent, derived from the high-resolution (1$^{\circ}$ $\times$ 1$^{\circ}$) daily gridded rainfall dataset provided by the India Meteorological Department (IMD) \cite{imd_rainfall_data}. From this daily data, we computed five distinct prediction targets: the mean rainfall (in mm/day) for each of the core monsoon months (June, July, August, September) and for the entire seasonal (JJAS) period. This formulation results in five independent forecasting tasks, and a separate model was trained for each target.

            Each of the five target datasets contains 85 years of 2D gridded rainfall fields, structured as a 4-dimensional tensor with a shape of \texttt{(Years, Latitude, Longitude, Channels)}, specifically \texttt{(85, 33, 35, 1)} for our study area. It is important to note that the IMD dataset primarily covers the Indian landmass; consequently, grid points falling outside this region are represented by 'NaN' values. To handle this, the 2D target grid for each year was transformed into a 1D target vector by selecting only the rainfall values at the 357 grid points corresponding to valid, non-NaN locations. This fixed-length vector serves as the ground truth for model implementation. After prediction, a reverse mapping process is used to reconstruct the full 2D gridded rainfall map, re-inserting the predicted values into their correct geographical coordinates and populating the remaining grid cells with NaNs for consistent visualization and spatial evaluation.            

            Each target dataset was split into training and testing sets using the exact same year-based division as the predictor data, ensuring a perfect one-to-one correspondence between each pre-monsoon input and its resulting monsoon rainfall output. To stabilize model training, the target rainfall values were also normalized using channel-wise Min-Max scaling. The normalization parameters were calculated from the training set targets and subsequently applied to both the training and testing sets, with any values in the test set falling outside the [0, 1] range being clipped.\\


        To the best of our knowledge, we are among the first to explicitly frame long-range monsoon prediction as a video-to-image problem, treating a rich set of 25 pre-monsoon variables as a multi-channel video. This allows the model to learn from the system's dynamics, not just its static state.

    \subsection{Evaluation Metrics}
    \label{ssec:metrics}
        To quantitatively assess model performance, we employ three regression metrics: Mean Squared Error (MSE), Mean Absolute Error (MAE), and the Sample-Normalized Mean Absolute Error (snMAE). These metrics are first computed over the valid target grid cells for each individual sample. The final reported score is the average of these values across all samples.
        
        \subsubsection{Mean Squared Error (MSE)}
            The MSE penalizes larger errors more heavily and is defined for a single sample as:
            \begin{equation}
            \label{eq:mse}
            \text{MSE} = \frac{1}{N} \sum_{i=1}^{N} \left( y_i - \hat{y}_i \right)^2,
            \end{equation}
            where $N$ is the number of valid grid points in the sample, and $y_i$ and $\hat{y}_i$ denote the ground truth and predicted values at the $i$-th valid grid cell, respectively.
        
        \subsubsection{Mean Absolute Error (MAE)}
            The MAE measures the average magnitude of absolute errors for a single sample:
            \begin{equation}
            \label{eq:mae}
            \text{MAE} = \frac{1}{N} \sum_{i=1}^{N} \left| y_i - \hat{y}_i \right|.
            \end{equation}
        
        \subsubsection{Sample-Normalized Mean Absolute Error (snMAE)}
            Standard Mean Absolute Percentage Error (MAPE) is ill-suited for rainfall prediction due to the frequent occurrence of zero-value grid points, which leads to undefined and unstable results. To create a stable, scale-independent metric, we define the snMAE. This metric computes the Mean Absolute Error (MAE) for each sample (i.e., each year's target map) and normalizes it by the mean of that target map. For a single sample, it is calculated as:
            \begin{equation}
            \label{eq:snmae}
            \text{snMAE} = \frac{\text{MAE}}{\mu + \epsilon} = \frac{\frac{1}{N} \sum_{i=1}^{N} |y_i - \hat{y}_i|}{\mu + \epsilon},
            \end{equation}
            where $\mu = \frac{1}{N} \sum_{i=1}^{N} y_i$ is the mean ground truth value for that sample, and $\epsilon$ is a small constant (e.g., $10^{-8}$) added for numerical stability. This approach provides a normalized error value without the risk of unstable results due to the frequent occurrence of zero-value grid points.

    \subsection{Model Framework}
    \label{ssec:architecture}
        Given the spatiotemporal nature of the input data and the limited number of training samples, we experimented deep 3D Convolutional Neural Networks (3D-CNNs) built upon residual connections. These architectures are well-suited for learning complex hierarchical features from sequence-to-image data while mitigating the vanishing gradient problem common in deep networks. The model follows a classic encoder-regressor design, which can be broken down into four main components: a spatiotemporal encoder, a temporal collapse layer, a spatial feature aggregator, and a final regression head.
        
        \subsubsection{3D Residual Block}
            The core building block of our network is a 3D residual block, inspired by the ResNet architecture \cite{he2016deep}. Each block consists of two 3D convolutional layers, with each convolution followed by Batch Normalization and a ReLU activation function. A shortcut connection adds the input of the block to its output after the second convolution, facilitating gradient flow. When the block performs downsampling (i.e., when strides are greater than one), a 1$\times$1$\times$1 convolutional projection is applied to the shortcut connection to match the new dimensions, ensuring compatibility for residual addition.
        
        \subsubsection{Spatiotemporal Encoder}
            The encoder consists of a stack of these 3D residual blocks. This component is responsible for progressively downsampling the input tensor in both spatial (latitude, longitude) and temporal (fortnights) dimensions. This hierarchical downsampling allows the model to learn features at multiple scales, from local patterns to larger, more abstract atmospheric structures.
        
        \subsubsection{Temporal Collapse and Feature Aggregation}
            Following the encoder, a crucial 3D convolutional layer is employed to collapse the temporal dimension entirely. This is achieved by setting the kernel's temporal size to match the full temporal extent of the feature map produced by the encoder. The output is a set of single, spatially-resolved 2D feature maps that aggregates information from all pre-monsoon fortnights. These 2D maps are then fed into a Global Average Pooling 2D (GAP) layer, which computes the mean of each feature channel across all spatial locations. The GAP layer produces a fixed-size feature vector that is robust to spatial translations and significantly reduces the number of model parameters compared to a simple flattening operation. The temporal collapse layer followed by the GAP layer also serves as a bottleneck that encourages the extraction of compact and high-level spatiotemporal features.
        
        \subsubsection{Regression Head}
            The final component is a regression head that maps the aggregated feature vector to the predicted rainfall vector. It consists of a dense hidden layer with 512 units and a ReLU activation, followed by a Dropout layer for regularization to prevent overfitting. The final output layer is a dense layer of 357 units with a sigmoid activation function.
        
        The model is compiled using the Adam optimizer \cite{Kingma2014AdamAM} and Mean Squared Error (MSE) as the loss function.
        
        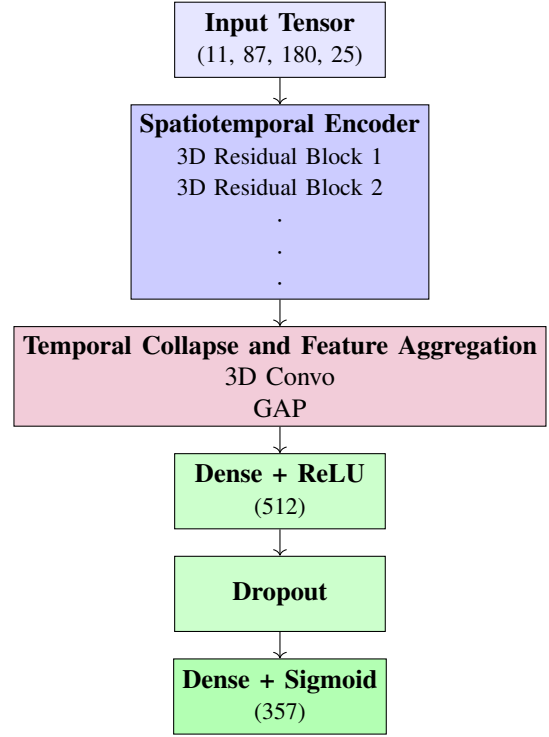
\begin{figure}[htbp]
            \centering
            \begin{tikzpicture}[node distance=1.4cm and 1cm, on grid, auto]
                \node[draw, minimum width=2.8cm, minimum height=1cm, fill=blue!10, align=center] (input) {\textbf{Input Tensor} \\ \small (11, 87, 180, 25)};
        
                \node[draw, minimum width=2.8cm, minimum height=2cm, below=1em of input.south, fill=blue!20, align=center] (encoder) {\textbf{Spatiotemporal Encoder}\\ \small 3D Residual Block 1\\\small 3D Residual Block 2\\ $\cdot$\\ $\cdot$\\ $\cdot$};
        
                \node[draw, minimum width=2.8cm, minimum height=1cm, below=1em of encoder.south, fill=purple!20, align=center] (collapse) {\textbf{Temporal Collapse and Feature Aggregation}\\ 3D Convo\\GAP};
        
                \node[draw, minimum width=2.8cm, minimum height=1cm, below=1em of collapse.south, fill=green!20, align=center] (dense1) {\textbf{Dense + ReLU} \\ \small (512)};
                \node[draw, minimum width=2.8cm, minimum height=1cm, below=1em of dense1.south, fill=green!20, align=center] (dropout) {\textbf{Dropout}};
                \node[draw, minimum width=2.8cm, minimum height=1cm, below=1em of dropout.south, fill=green!30, align=center] (output) {\textbf{Dense + Sigmoid} \\ \small (357)};
        
                \draw[->] (input) -- (encoder);
                \draw[->] (encoder) -- (collapse);
                \draw[->] (collapse) -- (dense1);
                \draw[->] (dense1) -- (dropout);
                \draw[->] (dropout) -- (output);
            \end{tikzpicture}
            \caption{Schematic diagram of the proposed 3D CNN-based model architecture.}
            \label{fig:model_architecture}
    \end{figure}

\section{Experiments and Results}
\label{sec:experiments}
    We conducted a comprehensive set of experiments to evaluate our framework, identify an optimal model configuration, and analyze its performance on an unseen test set.

    \subsection{Implementation and Setup}
        All experiments were conducted on a workstation equipped with an 18-core Intel(R) Xeon(R) W-2295 CPU @ 3.00GHz and 128GB RAM. The deep learning framework was implemented in Python using the TensorFlow and Keras libraries. As described in the methodology, Min-Max normalization was applied to scale all data to a [0, 1] range based on parameters derived exclusively from the training set. The normalization parameters for the five target variables, which are critical for denormalizing model output back to physical units (mm/day), are shown in Table \ref{tab:target_norm}.

        \begin{table}[h!]
        \centering
            \caption{Min-Max normalization parameters for target variables (rainfall in mm/day), derived from the training set.}
            \label{tab:target_norm}
            \resizebox{0.95\columnwidth}{!}{%
            \begin{tabular}{@{}lcc@{}}
                \toprule
                \textbf{Target Variable} & \textbf{Min Value (mm/day)} & \textbf{Max Value (mm/day)} \\ \midrule
                June & 0 & 58.0408 \\
                July & 0 & 71.1862 \\
                August & 0 & 91.0669 \\
                September & 0 & 43.9216 \\
                JJAS Average & 0 & 49.5925 \\
                \bottomrule
            \end{tabular}%
            }
        \end{table}

    \subsection{Quantitative Performance}
        Our experimental goal was not to claim a single best model, but rather to explore the viability of our proposed spatio-temporal framework for high-resolution forecasting. Due to the limited number of independent years available in the climate record (85 years total, 64 for training), we departed from a traditional train-validation-test split, as a further split would create a significantly small validation set. Instead, we trained a wide array of model configurations directly on the augmented training set and evaluated them on the held-out test set, relying on strong regularization (L2 and Dropout) and early stopping to prevent overfitting. This approach allows for a comprehensive analysis of the performance landscape across different architectural complexities.

        The test performance for different model configurations is summarized in Table \ref{tab:test_results}. 

        \begin{table}[h!]
        \centering
        \caption{Test performance for different model configurations.}
        \label{tab:test_results}
        \resizebox{\columnwidth}{!}
        {%
            \begin{tabular}{@{}cllcccc@{}}
                \toprule
                \textbf{Bottleneck Size} & \textbf{Target} & \textbf{Metric} & \textbf{1 Block} & \textbf{2 Blocks} & \textbf{3 Blocks} & \textbf{4 Blocks} \\ \midrule
                
                \multirow{15}{*}{64} & \multirow{3}{*}{June} & MSE & 0.00538 & 0.00538 & 0.00537 & 0.00537 \\
                 & & MAE & 0.04563 & 0.04572 & 0.04567 & 0.04574 \\
                 & & snMAE & 0.43878 & 0.44000 & 0.43934 & 0.44015 \\
                 
                & \multirow{3}{*}{July} & MSE & 0.00746 & 0.00743 & 0.00743 & 0.00743 \\
                 & & MAE & 0.05445 & 0.05448 & 0.05449 & 0.05449 \\
                 & & snMAE & 0.38244 & 0.38312 & 0.38323 & 0.38320 \\
                & \multirow{3}{*}{August} & MSE & 0.00302 & 0.00302 & 0.00302 & 0.00301 \\
                 & & MAE & 0.03889 & 0.03883 & 0.03886 & 0.03890 \\
                 & & snMAE & 0.41090 & 0.41027 & 0.41053 & 0.41104 \\
                 
                & \multirow{3}{*}{September} & MSE & 0.00997 & 0.00999 & 0.00994 & 0.00995 \\
                 & & MAE & 0.06901 & 0.06923 & 0.06868 & 0.06882 \\
                 & & snMAE & 0.55689 & 0.55898 & 0.55389 & 0.55518 \\
                & \multirow{3}{*}{JJAS Avg} & MSE & 0.00426 & 0.00426 & 0.00426 & 0.00425 \\
                 & & MAE & 0.04089 & 0.04090 & 0.04090 & 0.04090 \\
                 & & snMAE & 0.26452 & 0.26461 & 0.26462 & 0.26460 \\ \midrule

                 \multirow{15}{*}{128} & \multirow{3}{*}{June} & MSE & 0.00537 & 0.00537 & 0.00537 & 0.00537 \\
                 & & MAE & 0.04574 & 0.04567 & 0.04565 & 0.04565 \\
                 & & snMAE & 0.44028 & 0.43941 & 0.43922 & 0.43961 \\

                & \multirow{3}{*}{July} & MSE & 0.00745 & 0.00743 & 0.00744 & 0.00743 \\
                 & & MAE & 0.05447 & 0.05447 & 0.05449 & 0.05450 \\
                 & & snMAE & 0.38290 & 0.38334 & 0.38317 & 0.38333 \\

                & \multirow{3}{*}{August} & MSE & 0.00302 & 0.00302 & 0.00302 & 0.00302 \\
                 & & MAE & 0.03889 & 0.03887 & 0.03887 & 0.03888 \\
                 & & snMAE & 0.41094 & 0.41071 & 0.41068 & 0.41087 \\
   
                & \multirow{3}{*}{September} & MSE & 0.00995 & 0.00993 & 0.00995 & 0.00995 \\
                 & & MAE & 0.06880 & 0.06852 & 0.06868 & 0.06876 \\
                 & & snMAE & 0.55502 & 0.55243 & 0.55394 & 0.55461 \\
                       
                & \multirow{3}{*}{JJAS Avg} & MSE & 0.00427 & 0.00426 & 0.00425 & 0.00425 \\
                 & & MAE & 0.04083 & 0.04089 & 0.04091 & 0.04088 \\
                 & & snMAE & 0.26381 & 0.26381 & 0.26475 & 0.26448 \\ \midrule
   
                 \multirow{15}{*}{256} & \multirow{3}{*}{June} & MSE & 0.00536 & 0.00537 & 0.00537 & 0.00537 \\
                 & & MAE & 0.04591 & 0.04567 & 0.04573 & 0.04572 \\
                 & & snMAE & 0.44235 & 0.43940 & 0.44013 & 0.43996 \\

                & \multirow{3}{*}{July} & MSE & 0.00743 & 0.00744 & 0.00744 & 0.00743 \\
                 & & MAE & 0.05448 & 0.05447 & 0.05446 & 0.05447 \\
                 & & snMAE & 0.38314 & 0.38302 & 0.38294 & 0.38303 \\

                & \multirow{3}{*}{August} & MSE & 0.00302 & 0.00302 & 0.00302 & 0.00301 \\
                 & & MAE & 0.03890 & 0.03887 & 0.03886 & 0.03884 \\
                 & & snMAE & 0.41104 & 0.41074 & 0.41053 & 0.41032 \\

                & \multirow{3}{*}{September} & MSE & 0.00994 & 0.00997 & 0.00995 & 0.00995 \\
                 & & MAE & 0.06862 & 0.06893 & 0.06870 & 0.06880 \\
                 & & snMAE & 0.55336 & 0.55621 & 0.55403 & 0.55503 \\

                & \multirow{3}{*}{JJAS Avg} & MSE & 0.00426 & 0.00426 & 0.00426 & 0.00425 \\
                 & & MAE & 0.04093 & 0.04089 & 0.04089 & 0.04088 \\
                 & & snMAE & 0.26491 & 0.26451 & 0.26453 & 0.26448 \\ \midrule

                 \multirow{15}{*}{512} & \multirow{3}{*}{June} & MSE & 0.00556 & 0.00540 & 0.00540 & 0.00537 \\
                 & & MAE & 0.04538 & 0.04648 & 0.04558 & 0.04568 \\
                 & & snMAE & 0.43329 & 0.44763 & 0.43795 & 0.43946 \\

                & \multirow{3}{*}{July} & MSE & 0.00741 & 0.00744 & 0.00744 & 0.00743 \\
                 & & MAE & 0.05457 & 0.05448 & 0.05447 & 0.05447 \\
                 & & snMAE & 0.38454 & 0.38315 & 0.38301 & 0.38309 \\

                & \multirow{3}{*}{August} & MSE & 0.00302 & 0.00302 & 0.00301 & 0.00302 \\
                 & & MAE & 0.03895 & 0.03887 & 0.03886 & 0.03889 \\     
                 & & snMAE & 0.41166 & 0.41073 & 0.41057 & 0.41090 \\
                & \multirow{3}{*}{September} & MSE & 0.01032 & 0.00997 & 0.00995 & 0.00995 \\
                 & & MAE & 0.06997 & 0.06890 & 0.06874 & 0.06881 \\
                 & & snMAE & 0.56549 & 0.55598 & 0.55447 & 0.55513 \\ 
                & \multirow{3}{*}{JJAS Avg} & MSE & 0.00428 & 0.00516 & 0.00428 & 0.00426 \\
                 & & MAE & 0.04086 & 0.04847 & 0.04089 & 0.04088 \\
                 & & snMAE & 0.26417 & 0.31742 & 0.26453 & 0.26450 \\                 
                 \bottomrule
                 
            \end{tabular}%
            }
        \end{table}

    \subsection{Discussion of Results}
        A key finding from our experiments is the excellent generalization of the models. The results demonstrate robustness to architectural changes. The framework's performance is not highly sensitive to model complexity; it typically stabilizes after one or two blocks, with no significant benefit gained from deeper models. This leads to a crucial insight: a computationally efficient, simpler model is sufficient and practically preferable. This suggests our data representation is effective at extracting the available predictive signal, and the primary limiting factor is likely the inherent climatic predictability rather than model capacity.

        To understand the practical implication of the error metrics, we can interpret, for example, the MAE in its physical units. The physical MAE, $\text{MAE}_{\text{phys}}$, can be calculated from the normalized MAE, $\text{MAE}_{\text{norm}}$, using the scaling parameters from Table \ref{tab:target_norm}.
        \begin{equation}
        \label{eq:mae_denorm}
        \text{MAE}_{\text{phys}} = \text{MAE}_{\text{norm}} \times (y_{\text{max}} - y_{\text{min}})
        \end{equation}
        This value represents the average absolute deviation of the model's prediction from the ground truth in mm/day. To understand the practical implications of the error metrics, we can denormalize them back into their physical units (mm/day) using the test results from a representative model (1-block, 64 bottleneck size, June target). The MAE provides the most direct measure of average prediction error. A normalized MAE of 0.04563 corresponds to a physical error of $0.04563 \times (58.0408 - 0) \approx 2.65 \text{ mm/day}$, meaning a typical grid-point prediction deviates from the truth by this amount.        
        
\begin{figure}[h!]
          \centering
          \includegraphics[width=\columnwidth]{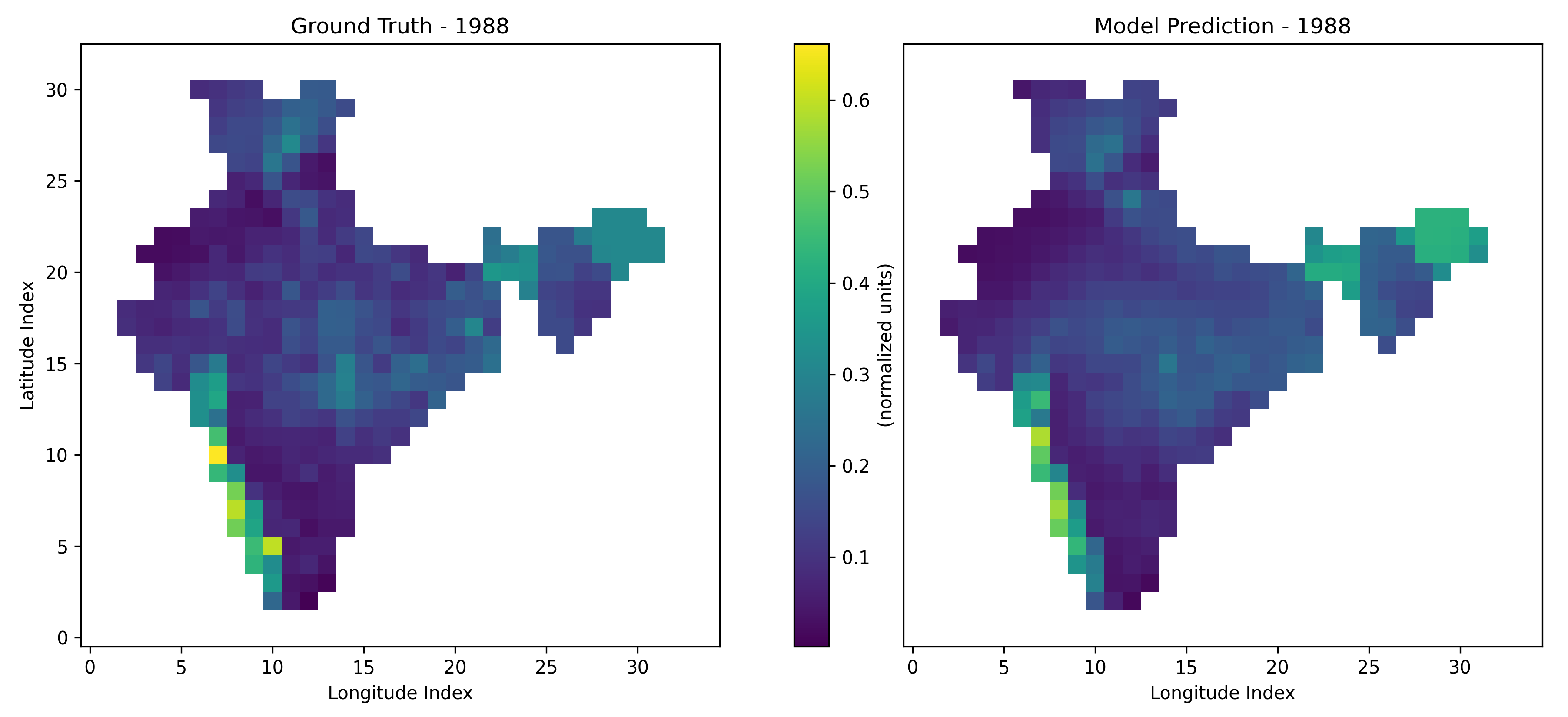}
          \caption{Example of a strong prediction for the JJAS seasonal average (Year 1988, Bottleneck size- 512, 4 Blocks). The model (right) closely predicts the spatial patterns of the ground truth (left).}
          \label{fig:good_prediction}
        \end{figure}

\begin{figure}[h!]
          \centering
          \includegraphics[width=\columnwidth]{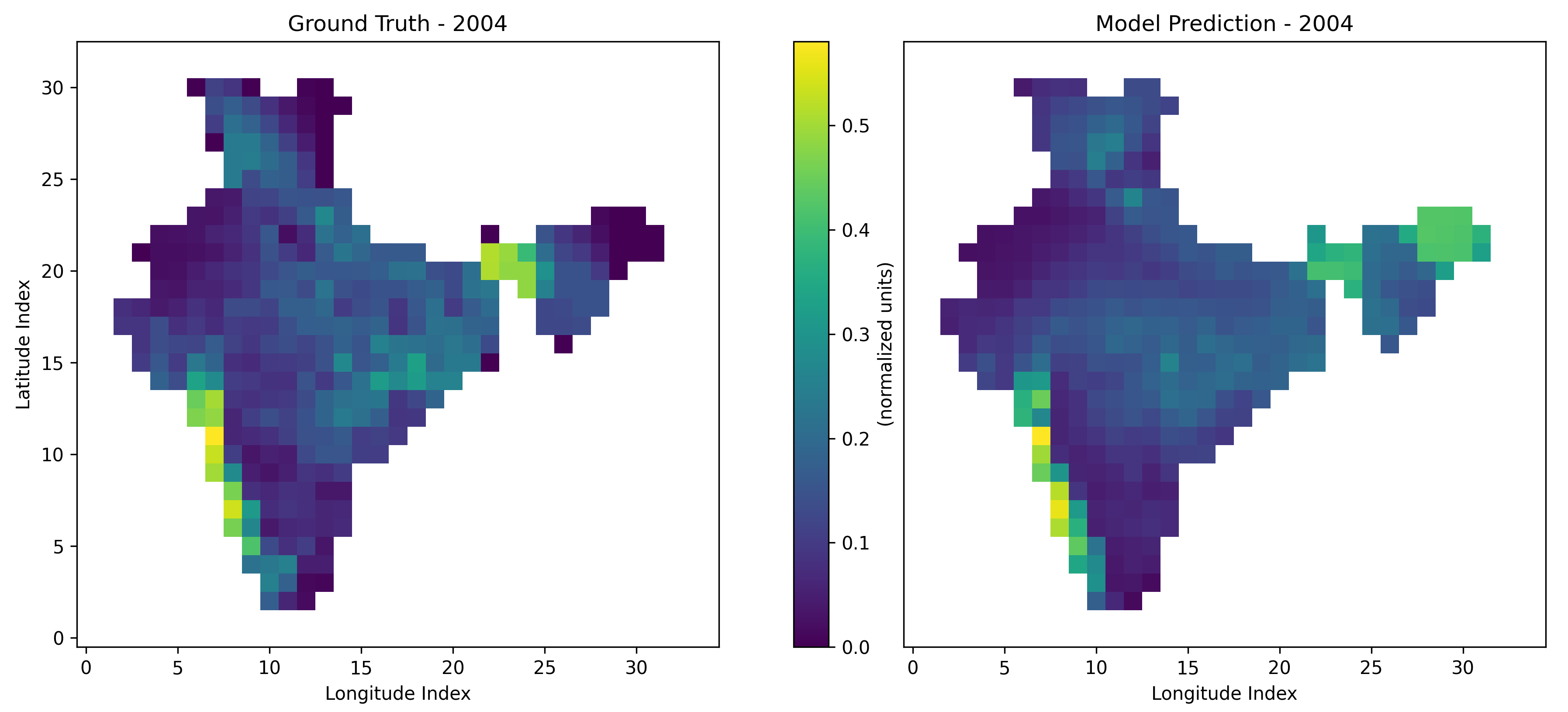}
          \caption{Example of a weaker prediction for the JJAS seasonal average (Year 2004, Bottleneck size- 512, 4 Blocks). The model (right) fails to capture the high-intensity rainfall in the Northeast India seen in the ground truth (left).}
          \label{fig:bad_prediction}
        \end{figure}

        For a visual assessment of spatial prediction, we present two representative cases from the seasonal (JJAS) average test set. Figure \ref{fig:good_prediction} exemplifies a year of strong predictive performance. In this case, the model successfully captures the large-scale spatial distribution of rainfall, efficiently identifying the high-rainfall regions along the Western Ghats and in Northeast India, as well as the drier zones in the rain-shadow region of central India. The overall pattern correlation is high, demonstrating the model's ability to learn meaningful climatic relationships. Conversely, Figure \ref{fig:bad_prediction} illustrates a case where the model's performance is more limited. While the broad pattern is still recognizable, the model significantly underestimates the peak rainfall in the north-eastern states. This discrepancy, where the model struggles with regions of extreme rainfall, is a consistent challenge observed across a few years in the test set.
        
        This geographical heterogeneity in performance is likely attributable to the extreme spatio--temporal variability of rainfall in these orographically complex regions. The available data, while extensive chronologically, may be insufficient to fully resolve the intricate local dynamics governing these high-intensity events.        


\section{Conclusion}
\label{sec:conclusion}
    
    This paper introduced a deep learning framework that reframes long-range monsoon forecasting as a computer vision problem. By treating multi-variable pre-monsoon data as a video-like input, our CNN-based model successfully produces high-resolution, gridded rainfall forecasts for both individual monsoon months and the seasonal average. Results demonstrate that this approach is robust, with even compact models generalizing effectively to unseen data. The ability to generate spatially detailed, intra-seasonal predictions is a significant advance beyond traditional single-index forecasts, offering more actionable information for regional water and agricultural management. While our framework provides a strong baseline, performance is sometimes limited in regions with extreme orographic rainfall. Key directions for future work include developing probabilistic forecasts to quantify uncertainty, exploring advanced architectures such as Vision Transformers, exploring Generative AI for data augmentation and applying explainable AI techniques to uncover new scientific insights into monsoon dynamics.

\section*{Acknowledgments}
    This work was supported by the Startup Research Grant (SRG) of the Science and Engineering Research Board (SERB), which is now a part of the Anusandhan National Research Foundation (ANRF), under the Department of Science and Technology, Government of India.

\bibliography{bib_file}

\vfill

\end{document}